\documentclass[10pt]{article}

\usepackage{amsmath,graphicx}
\usepackage[T1]{fontenc}
\usepackage[utf8]{inputenc}
\usepackage{tabularx}
\usepackage{url}
\DeclareGraphicsExtensions{.pdf,.jpeg,.png,.tif,.eps,.jpg}

\DeclareMathOperator{\efunction}{e}
\newcommand{\expo}[1]{{\efunction^{#1}}}

\newcommand{\absval}[1]{{\left|#1\right|}}
\DeclareMathOperator{\arctanh}{arctanh}

\begin{document}
\title{Saccadic Eye Movements and the Generalized Pareto Distribution}

\author{Reiner Lenz\\
Dept. Science and Technology\\
Linköping University\\
SE-60174 Norrköping\\
Sweden\\
Email: reiner.lenz@liu.se}
\date{}
\maketitle
\begin{abstract}
We describe a statistical analysis of the eye tracker measurements in a database with 15 observers viewing 1003 images under free-viewing conditions. In contrast to the common approach of investigating the properties of the fixation points we analyze the properties of the transition phases between fixations. We introduce hyperbolic geometry as a tool to measure the step length between consecutive eye positions. We show that the step lengths, measured in hyperbolic and euclidean geometry, follow a generalized Pareto distribution. The results based on the hyperbolic distance are more robust than those based on euclidean geometry. We show how the structure of the space of generalized Pareto distributions can be used to characterize and identify individual observers.  
\end{abstract}
\subsection*{Keywords}
Eye tracking, generalized pareto distribution, hyperbolic geometry, distribution fitting, classification

\section{Introduction}

The investigation of eye movements is important in the understanding of basic vision properties and with the availability of new eye tracker technology it is also of growing practical importance. Typical research topics are the relation between image properties and eye movements (bottom-up), the influence of high-level goals on movement patterns (top-down) or models to predict salient regions in an image (for an overview see~\cite{Borji2013}). In this study we use a model which describes the location of the eye positions as the realization of a stochastic process that consists of two components, one component that describes the larger, jump-type changes and a second component related to the relatively small changes of the eye positions. This saccade-and-fixate strategy is one of the fundamental processes used by the human visual system to analyze its environment. 

The factors that control these processes are very complex. They include high-level task-solving factors and input driven factors depending on the visual properties of the current input image. In the following we will ignore all these factors and we will only analyze the statistical properties of eye-tracker data. The problem we want to investigate is: can we characterize individual observers from a collection of eye-tracker measurements? Furthermore we will ignore fixation points, which are perhaps more controlled by task and/or image related factors, and we will only use parameters derived from the saccadic movements. We investigate this problem with the help of a large database where 15~observers viewed 1003~images in free-viewing conditions~\cite{Tilke2009}. 

The major steps in this investigation are the following: First we describe two methods to compute the step-length between two different eye positions. The first is the usual euclidean length while the second is using the disc version of hyperbolic geometry which takes into account that the viewing space of the observer is a cone. Transition points between fixations are by definition points with comparatively large distances between consecutive eye positions. If we consider saccadic eye movements as a stochastic process then these non-fixation points correspond to the tail of the distribution of step-lengths. Such tail distributions follow very often the generalized Pareto distribution (GPD). In the second step of our analysis we will show that the distribution of the step-lengths of the saccadic eye movements can indeed be described by the GPD. The GPD depends on three parameters: position~($\theta$), shape~($k$) and scale~($\sigma$). In the third part we choose a random selection of images and for each user we fit the GPD to the data. Each such experiment results in a 3-D parameter vector for each user. Selecting only a few images will obviously lead to results where the variation between images is larger than the variation between users. We will however show that with an increasing number of images per trial clear clusters appear in the parameter space. These clusters can be characterized with a mixture-of-Gaussian model. Distances in parameter space are a poor way to characterize the similarity between probability distributions but the structure in the parameter space suggests that it should be possible to identify individual observers based on the distribution of their saccadic step lengths. In our final experiments we therefore use samples of the distributions as feature vectors and we train support-vector machines (SVM) to discriminate between one specific observer and the rest. These experiments result in very high recognition rates.
\section{Material and methods}

The data used in this study is described in~\cite{Tilke2009}. Together with useful code it can be downloaded from the website of the authors\footnote{~\url{http://people.csail.mit.edu/tjudd/WherePeopleLook/index.html}}. The database  contains eye tracking data of 15~viewers and 1003~images. The database contains also code to compute fixation points. In the following we use this code to identify fixation points and to extract those eye tracking measurements that are related to the saccadic movements between fixations. 

\begin{figure}
	\centering
		\includegraphics[width=0.5\columnwidth]{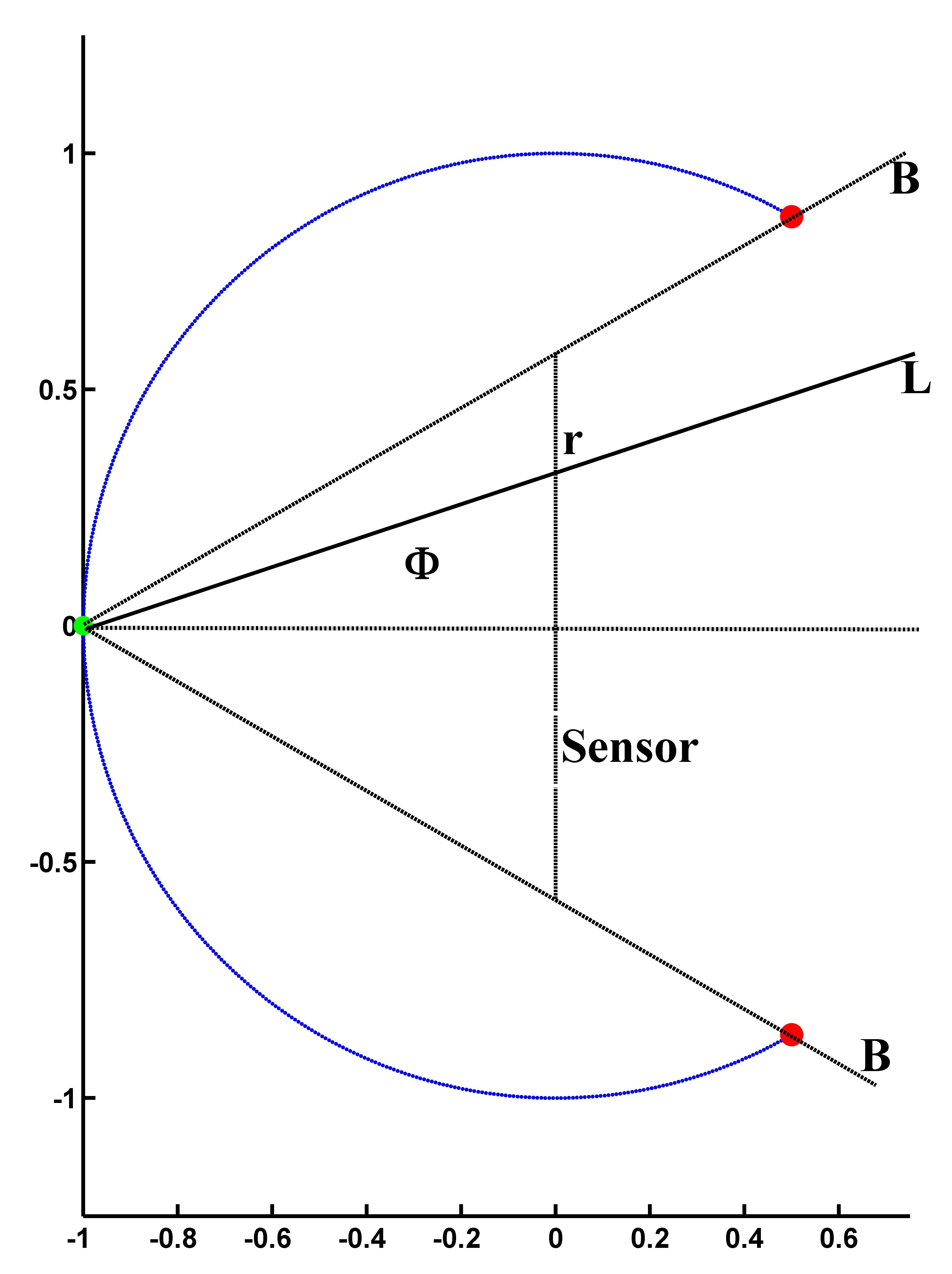}
	\caption{Hyperbolic}
	\label{fig:PinholeCamera}
\end{figure}

We ignore the direction of the movements and use only their step lengths. The eye-positions in the database are given in a planar coordinate system. A natural way to compute the distance between two position vectors is thus given by the common euclidean distance. In the following we will also use a model based on hyperbolic geometry. We don't go into the technical details (which can be found in books on non-euclidean geometry~\cite{Anderson1999}, \cite{Siegel1969} has a short introduction) but we will give an intuitive motivation. Consider the model of a pinhole camera in Figure~\ref{fig:PinholeCamera}. The field of view of the camera is restricted to the cone between the two lines denoted by~{\bfseries B}. A point in the scene located somewhere along the line~{\bfseries L} is mapped to the pixel~{\bfseries r} on the sensor. All projection lines go through the 'pinhole' located at position~$(-1,0)$ (note that it is also possible to place the sensor behind the pinhole which gives essentially the same geometry). In euclidean geometry based models of eye movements one often uses two angles to describe the motion of the eye, see~\cite{Haslwanter1995}. A corresponding model for the one-dimensional model in Figure~\ref{fig:PinholeCamera} would use the angle~$\Phi$ to characterize the line~{\bfseries L}. In this framework there is no build-in mechanism that requires that the projection line~{\bfseries L} lies between the two lines denoted by~{\bfseries B}. If we use the sensor coordinate~{\bfseries r} then we can require that the absolute value of~{\bfseries r} lies between zero and one and we can use the new, hyperbolic angle~$\rho$ defined as~$\absval{r} = \tanh{\rho}$ as a coordinate for~{\bfseries L}. For points near the origin the hyperbolic distance is similar to the euclidean distance by for points near the boundary it goes to infinity. Note also that instead of the euclidean distance~{\bfseries r} with the rather arbitrary upper bound of one we can use the hyperbolic angle~$\rho$ as a distance to the origin and then we have a natural distance measure with the only restriction that it should be non-negative.
In the case of a three-dimensional scene and a two-dimensional sensor, the sensor is given by the unit disk. The points on the sensor can be described  by euclidean coordinates~$(x,y),$ but it is easier to use complex coordinates~$z = x + iy = \tanh(\rho)\expo{i\varphi}$ where~$\rho$ is defined as in the one-dimensional case. For two sensor points~$z = x+iy, w = u+iv$ their hyperbolic distance is given by 
\[
\arctanh\absval{\frac{z-w}{1-z\overline{w}}}
\]
which reduces to the one-dimensional definition in the case where~$w=0$ and~$z$ real and~$0\leq z\leq1$.

For pairs of consecutive points we compute the distance between them and consider the statistical properties of these step lengths. In the terminology of random walks we consider the eye-tracking data as a random walk where the step lengths follow a heavy tailed distribution. Furthermore we are only interested in those parts of the walk in which the step lengths are large. Here we do not define a numerical value of that threshold but we simply restrict our investigation to points that are not classified as fixation points in the database. Data of this type is often investigated with the help of Pareto distributions that are used when excesses of a random variable over a threshold are considered. The probability density function (PDF) of the generalized Pareto distribution is defined as: 
\[p(x;k,\sigma,\theta) = \left\{
\begin{array}{cc}
 \ & 
\begin{array}{cc}
 \frac{\left(1+\frac{k  (x-\theta )}{\sigma }\right)^{-\frac{1}{k }-1}}{\sigma } & \theta <x<\left(
\begin{array}{cc}
 \{ & 
\begin{array}{cc}
 \theta -\frac{\sigma }{k } & k <0 \\
 \infty  & \text{otherwise} \\
\end{array} 
 \\
\end{array}
\right) \\
 0 & \text{otherwise} \\
\end{array}
 \\ 
\end{array}
\right.
\]
We estimate the parameters of the distribution with the help of the Matlab function gpfit and therefore we follow the notation used there: $k$~is the shape, $\sigma$~is the scale and~$\theta$ is the location parameter of the distribution. Note that the shape parameter~$k$ can be negative in which case the support of the distribution is a finite interval. For positive~$k$ the support is the positive half-axis with left end point at~$\theta.$ We do not consider the special case of~$k=0$. More information about the generalized Pareto distribution and extreme value distributions in general can be found in~\cite{Castillo2005} (but note the sign change for~$k$ there). The estimation of the parameters is done in two steps: first we find the minimum distance value in the data vector. Then we shift the distance values so that the minimum value is zero. In the second step we ignore all shifted distance values with value exactly equal to zero and fit a two parameter GPD with shape and scale parameters to the shifted data. This gives a three parameter representation of the data in terms of the minimum value and the shape and scale parameters of the GPD. 

Estimating distribution parameters for single images and a single user is obviously not very meaningful. One problem is that the number of non-fixation points is, by definition, much lower than the number of fixation points and the other reason is that the form of a given eye movement sequence can vary considerably. In our experiments we therefore select randomly a given number of images from the database. For a given observer we combine all non-fixation measurements from the corresponding observations into one dataset. From this dataset we estimate the parameters of the GPD. In most experiments described below this process is repeated 5000 times to see how the random selection of the images influences the values of the estimated parameters. The results show that the distribution parameters are concentrated in clusters linked to the different observers. 

In the final experiments we use samples from the probability density functions of the GPD's and train support-vector machines (SVM) to discriminate between one observer and the remaining observers. We will show that the recognition rates are very high and that they are slightly better in the euclidean metric based experiments than in the experiments using the hypberbolic distances.

\section{Experiments and Results}
\begin{figure}
	\centering
		\includegraphics[width=0.75\columnwidth]{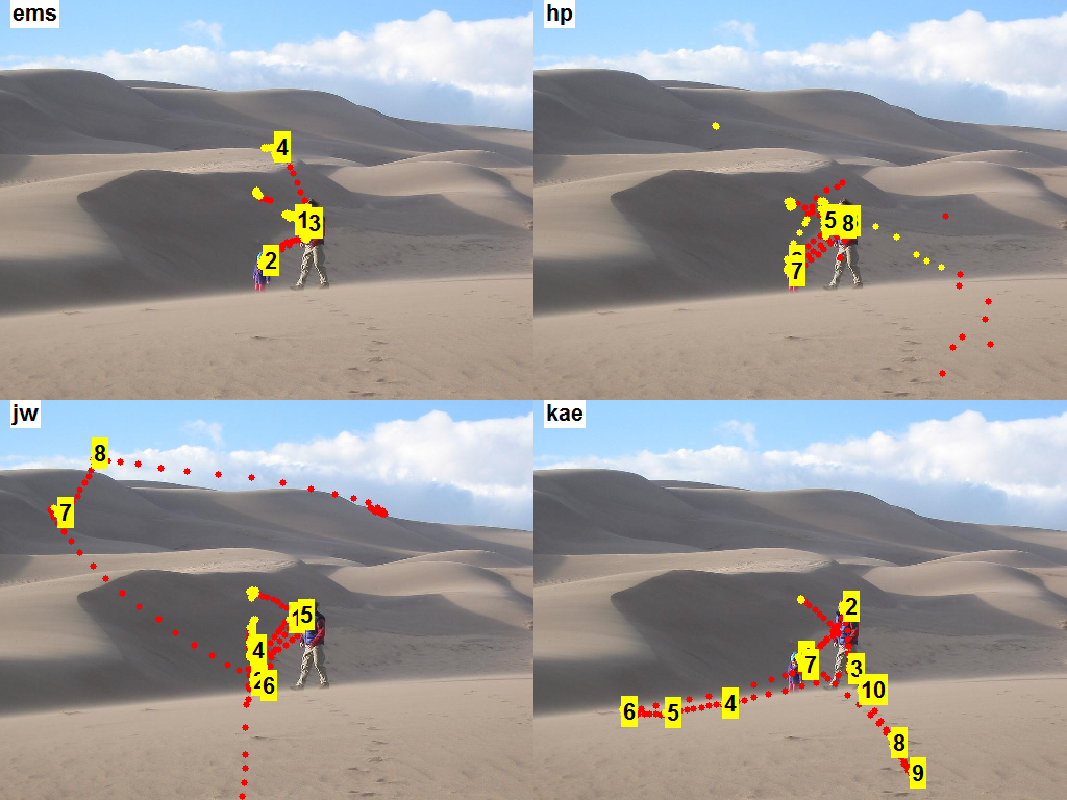}
	\caption{Image i14020903.jpeg, Viewers: ems,hp,jw,kae}
	\label{fig:Montage_i14020903}
\end{figure}

\begin{figure}
	\centering
		\includegraphics[width=0.75\columnwidth]{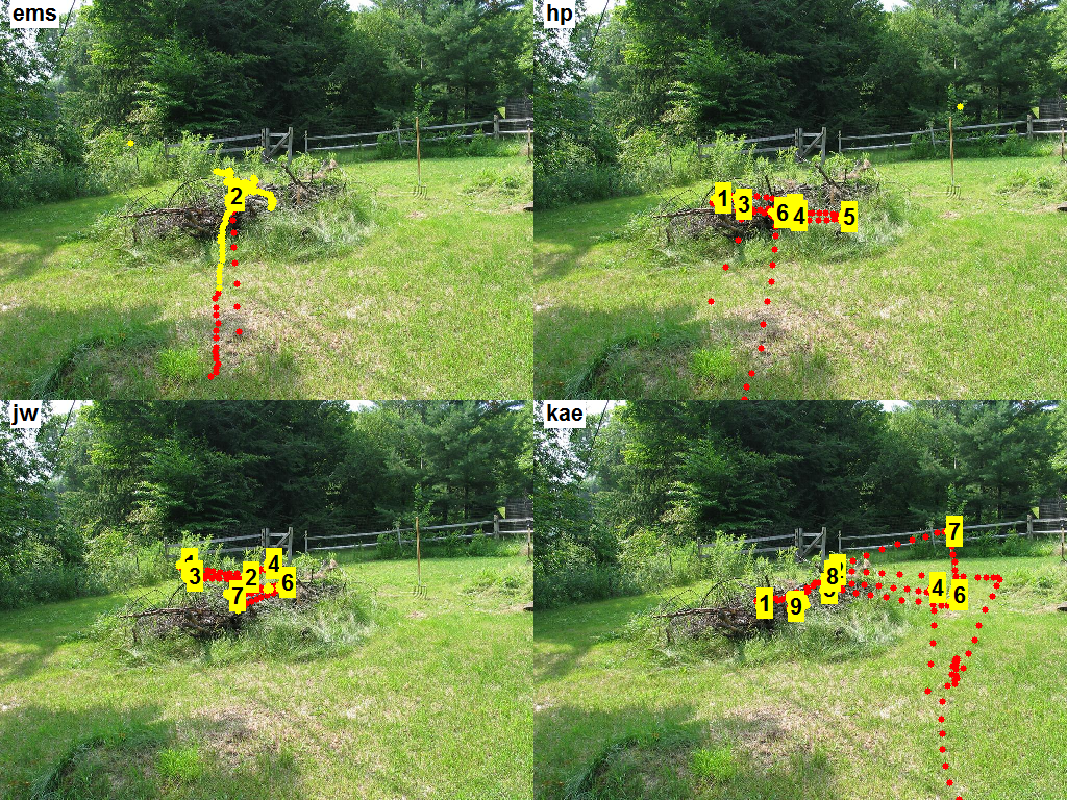}
	\caption{Image i113347896.jpeg, Viewers: ems,hp,jw,kae}
	\label{fig:Montage_i113347896}
\end{figure}

As an illustration of the variation of the eyetracking measurements for different observers we show in Figures~\ref{fig:Montage_i14020903} and~\ref{fig:Montage_i113347896} the measurements for four observers ({\it ems, hp, jw, kae}) when viewing the two images ({\it i14020903.jpeg, i113347896.jpg}) in the database.
    
\begin{figure}
	\centering
		\includegraphics[width=0.75\columnwidth]{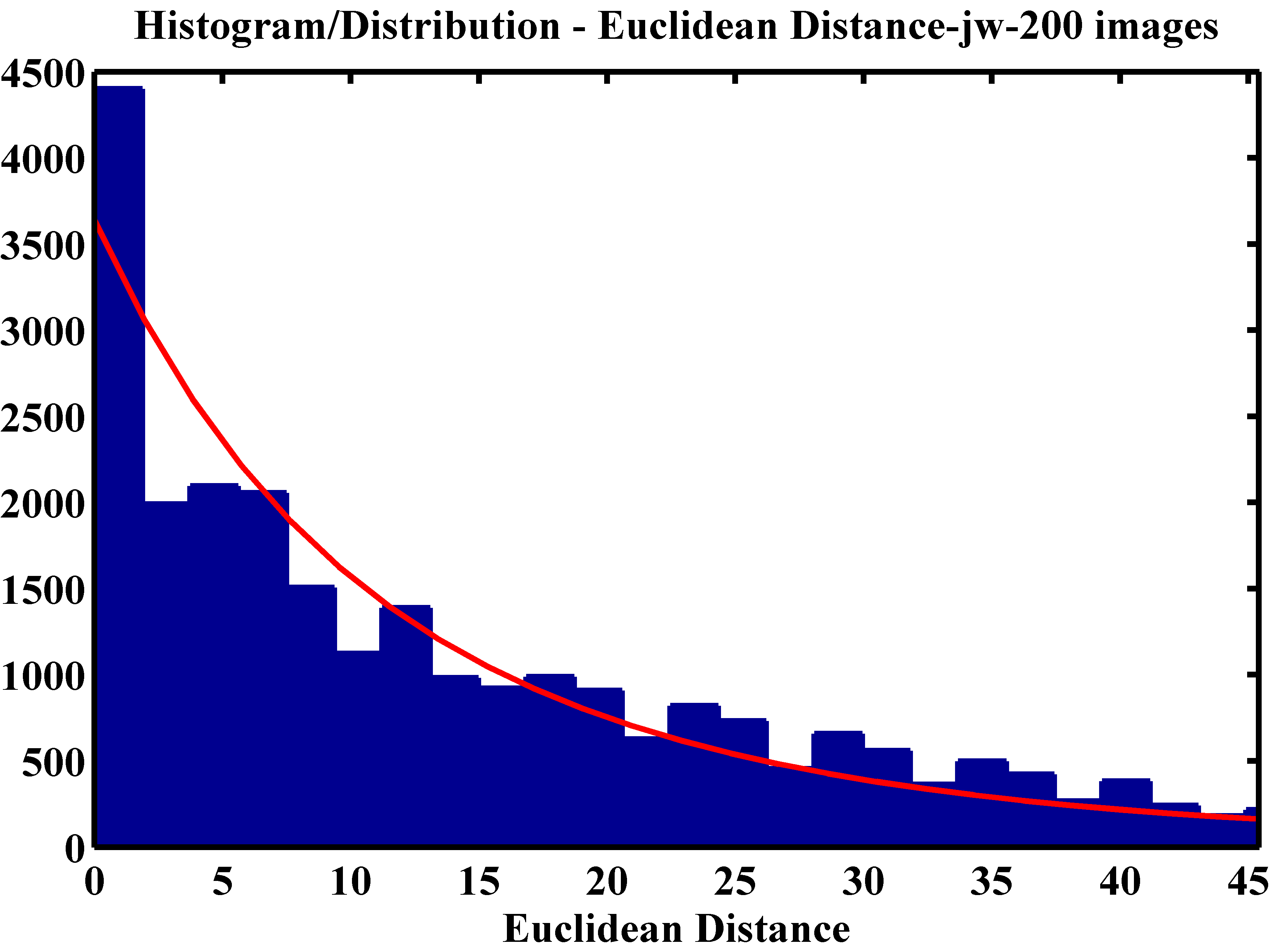}
	\caption{Histogram and GPD using euclidean distance, 200 images, observer jw}
	\label{fig:EucHistDistJW200}
\end{figure}

\begin{figure}
	\centering
		\includegraphics[width=0.75\columnwidth]{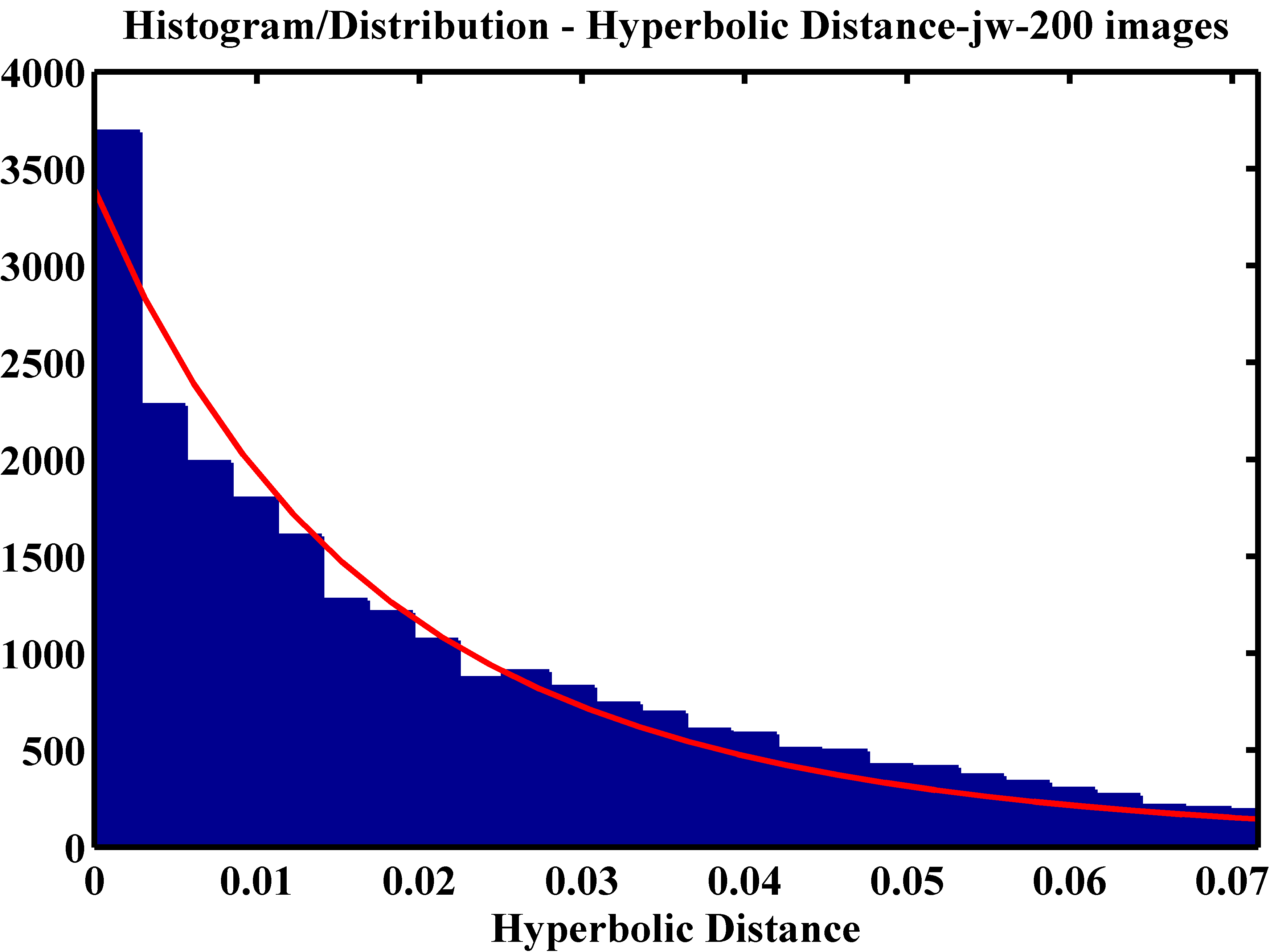}
	\caption{Histogram and GPD using hyperbolic distance, 200 images, observer jw}
	\label{fig:HypHistDistJW200}
\end{figure}

\begin{figure}
	\centering
		\includegraphics[width=0.75\columnwidth]{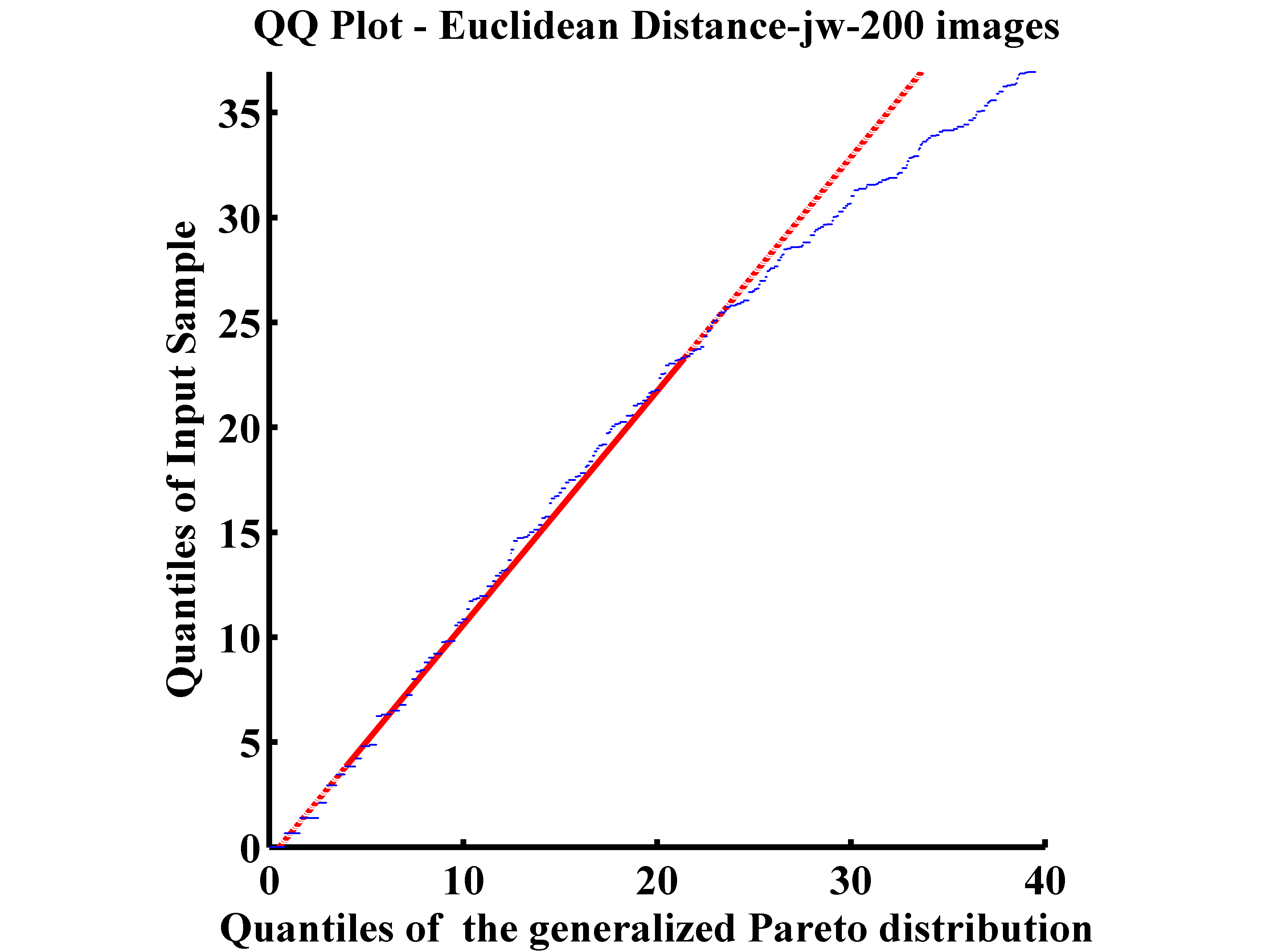}
	\caption{Quantile-quantile plot using euclidean distance, 200 images, observer jw}
	\label{fig:EucQQPlotJW200}
\end{figure}

\begin{figure}
	\centering
		\includegraphics[width=0.75\columnwidth]{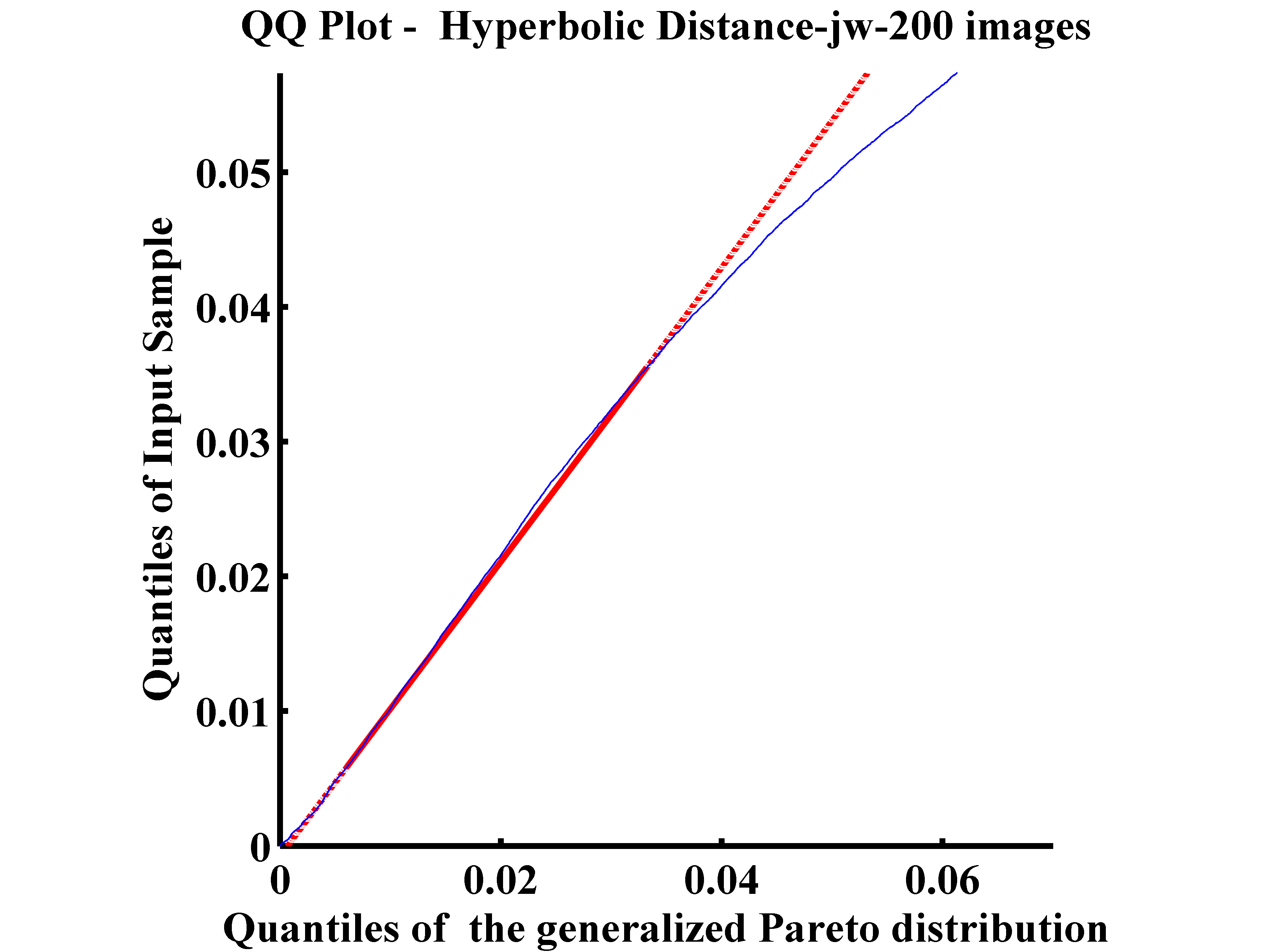}
	\caption{Quantile-quantile plot using hyperbolic distance, 200 images, observer jw}
	\label{fig:HypQQPlotJW200}
\end{figure}

After fitting the GPD to the data we computed the adjusted R-squared value (see~\cite{Upton2008}) which gives a measure of how similar the shapes of the empirical and the fitted distribution functions are. These values lie between zero and one and higher values indicate better fit. In the following experiments we selected 50, 100, 200 and 500 random images in each trial and we fitted the GPD for each observer. We did this for both the euclidean and the hyperbolic distances. The mean values of the adjusted R-squared values are collected in Table~\ref{tab:em} for the euclidean distance and in Table~\ref{tab:hm} for the hyperbolic distance. The column {\it All} contains the mean value over all observers. From the tables one can see that the hyperbolic distance results are in general slightly better and they produce more consistent fitting results. As an illustration of what these numbers really mean, we compare the empirical distribution and the estimated GPD in Figures~\ref{fig:EucHistDistJW200}, \ref{fig:HypHistDistJW200},\ref{fig:EucQQPlotJW200} and~\ref{fig:HypQQPlotJW200}. In the experiment we selected 200~random images and the eye-tracker data of the observer~{\it jw}. The first two plots show the histogram of the distribution and the fitted GPD, both for the euclidean and the hyperbolic distances. These distributions have long tails and therefore we restrict the plot range in Figures~\ref{fig:EucHistDistJW200}, \ref{fig:HypHistDistJW200} to the relevant parts of the distributions. Figures~\ref{fig:EucQQPlotJW200} and~\ref{fig:HypQQPlotJW200} are quantile-quantile plots of the same data. These plots show the relations between the quantiles of the empirical data and the corresponding quantiles of the GPD. For a perfect fit all points lie on the 45 degrees diagonal. We see again that the fit is very good for the major parts of the distributions, only for very high quantiles the differences become noticeable. Which is natural in this case since we have, by definition, very few observations in this value range.

\begin{table}
	\centering
		\begin{tabularx}{\textwidth}{r||c | c | c | c | c | c | c | c}
			Im.&All&CNG&ajs&emb&ems&ff&hp&jcw\\
			\hline
 50 &0.9912 & 0.9920 & 0.9917 & 0.9909 & 0.9951 & 0.9922 & 0.9862 & 0.9952\\
100 &0.9912 & 0.9921 & 0.9917 & 0.9909 & 0.9958 & 0.9918 & 0.9859 & 0.9955\\
200 &0.9910 & 0.9919 & 0.9916 & 0.9907 & 0.9962 & 0.9913 & 0.9854 & 0.9956\\
500 &0.9906 & 0.9915 & 0.9913 & 0.9905 & 0.9964 & 0.9907 & 0.9847 & 0.9956\\
\hline
	Im.&	 jw & kae & krl & po & tmj & tu & ya & zb\\
\hline
  50 &0.9877 & 0.9941 & 0.9853 & 0.9963 & 0.9942 & 0.9919 & 0.9898 & 0.9851\\
 100 &0.9875 & 0.9946 & 0.9849 & 0.9965 & 0.9944 & 0.9916 & 0.9899 & 0.9844\\
 200 &0.9872 & 0.9949 & 0.9844 & 0.9965 & 0.9946 & 0.9911 & 0.9899 & 0.9835\\
 500 &0.9869 & 0.9952 & 0.9837 & 0.9964 & 0.9945 & 0.9904 & 0.9896 & 0.9824
\end{tabularx}
	\caption{Mean adjusted R-squared values for the euclidean distances}
	\label{tab:em}
\end{table}

\begin{table}
	\centering
		\begin{tabularx}{\textwidth}{r||c | c | c | c | c | c | c | c}
			Im. &All  & CNG & ajs & emb & ems & ff & hp & jcw\\
			\hline
 50&0.9958 & 0.9953 & 0.9956 & 0.9965 & 0.9961 & 0.9931 & 0.9959 & 0.9978\\
100&0.9960 & 0.9955 & 0.9959 & 0.9967 & 0.9964 & 0.9932 & 0.9963 & 0.9979\\
200&0.9961 & 0.9956 & 0.9960 & 0.9967 & 0.9966 & 0.9932 & 0.9965 & 0.9980\\
500&0.9962 & 0.9957 & 0.9960 & 0.9968 & 0.9966 & 0.9932 & 0.9966 & 0.9981\\
\hline
		Im.&	 jw & kae & krl & po & tmj & tu & ya & zb\\
			\hline
 50&0.9954 & 0.9965 & 0.9943 & 0.9970 & 0.9979 & 0.9974 & 0.9953 & 0.9934\\
100&0.9955 & 0.9966 & 0.9945 & 0.9971 & 0.9981 & 0.9977 & 0.9955 & 0.9936\\
200&0.9956 & 0.9967 & 0.9946 & 0.9971 & 0.9982 & 0.9978 & 0.9956 & 0.9938\\
500&0.9957 & 0.9968 & 0.9946 & 0.9971 & 0.9983 & 0.9978 & 0.9957 & 0.9938
\end{tabularx}
	\caption{Mean adjusted R-squared values for the hyperbolic distances}
	\label{tab:hm}
\end{table}

\begin{figure}[htbp]
	\centering
		\includegraphics[width=0.75\columnwidth]{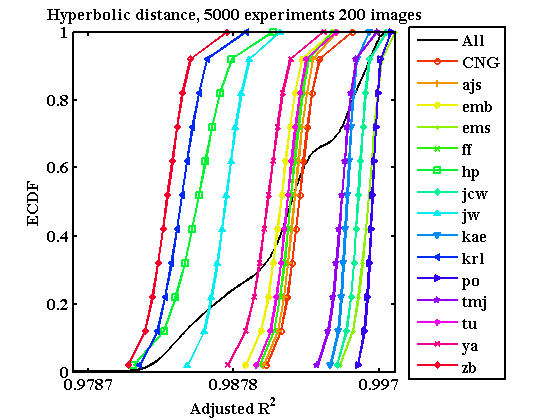}
	\caption{Distribution of $R^2$-parameters for 5000 experiments based on 200 images each}
	\label{fig:Pareto_5000_200_RSqrd}
\end{figure}

Mean values of the adjusted R-squared values give only a summary overview over the computed values. In Figure~\ref{fig:Pareto_5000_200_RSqrd} we show for all the observers the empirical cumulative distribution function (ECDF) of these values for the case where we used the hyperbolic distance, computed 5000~fittings and used 200~images in each run. 

The distribution of the data points are described by three parameters: the minimum value, the shape and the scale parameters. The experiments confirmed that the minimum values are indeed very small and the variance of the minimum values is also almost zero (typically of the order $10^{-10}$). The following description of the structure of the parameter space of the distributions is therefore restricted to the distribution of the shape~$k$ and scale parameters~$\sigma$ of the GPD, fitted to the shifted distance values. In the classification experiments we will make use of the three-parameter form of the GPD.

An analysis of the distribution of the ($k,\sigma$)-parameters showed that the ($k,\sigma$)-vectors for a given observer are concentrated in restricted regions and we therefore fitted Gaussian mixture models (with 15 Gaussians, using the Matlab function gmdistribution.fit) to the distribution of the ($k,\sigma$)-vectors.

\begin{figure}[htbp]
	\centering
		\includegraphics[width=0.75\columnwidth]{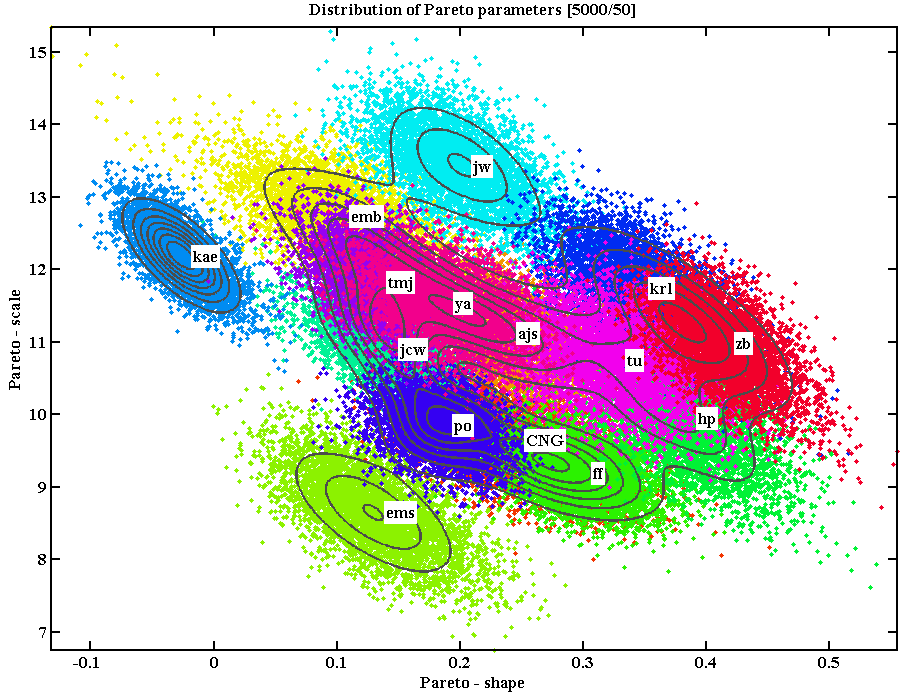}
	\caption{Distribution of shape-scale parameters for 5000 trials and 50 images per trial}
	\label{fig:Pareto_5000_50_ksigma}
\end{figure}

\begin{figure}[htbp]
	\centering
		\includegraphics[width=0.75\columnwidth]{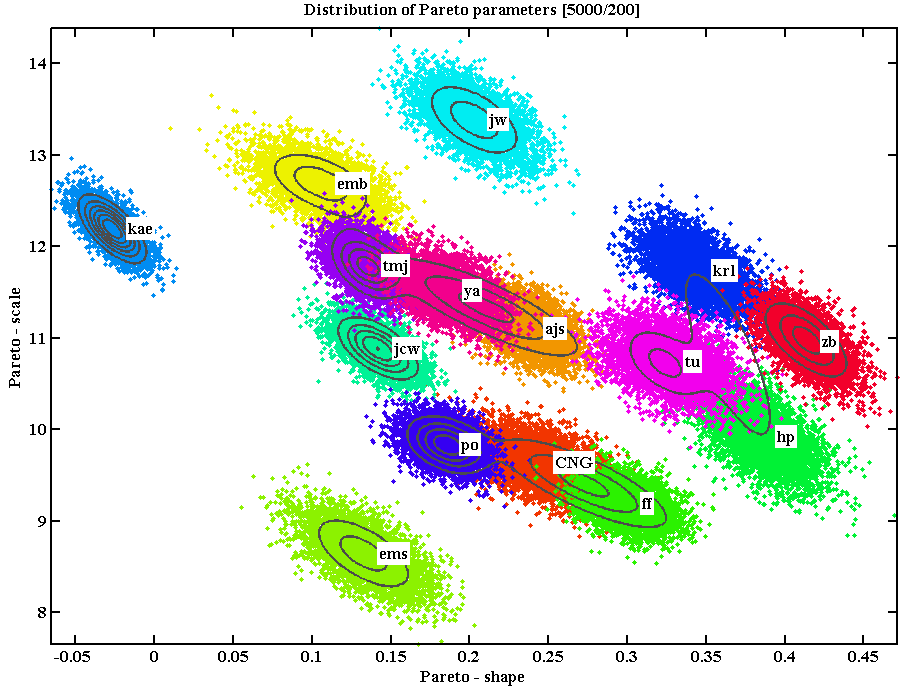}
	\caption{Distribution of shape-scale parameters for 5000 trials and 200 images per trial}
	\label{fig:Pareto_5000_200_ksigma}
\end{figure}

For the experiments where 50, 
and 200~images per trial were used in 5000 trials we show the distribution of the GPD parameters and the overlayed Gaussian mixture contours in Figures~\ref{fig:Pareto_5000_50_ksigma}, 
and~\ref{fig:Pareto_5000_200_ksigma}. The hyperbolic distance is used in all figures. The abbreviated user name is displayed at the position of the value of the median of the shape parameter~$k$ and the scale parameter~$\sigma$  for that user.We see that by increasing the number of images considered in each trial the separation between the users improves and for 50
and 200 images the 15 Gaussians give a good description of the 15 observers. One interesting observation is the location of the parameters for observer {\it kae} where fitting resulted in a negative shape parameter. Note that the observers {\it ems, hp, jw, kae} used in Figure~\ref{fig:Montage_i14020903} and Figure~\ref{fig:Montage_i113347896} are those that occupy the outlying regions in the ($k,\sigma$)-space.

The structure in the distribution parameter space indicates that it should be possible to identify individual observers from the distribution of their saccadic eye movements. 
In a series of experiments we therefore investigated if this really is the case. 

First we note that the distributions are characterized by the three parameters from which the complete pdf can be computed. In the case where we used $L$~trials in the fitting experiment the whole information can be stored in a database of size~$L\times 15\times 3$ (number of trials x number of observers x number of distribution parameters). Such databases will be used in the following classification experiments, the values for $L$ are 5000 and 10000.

The distributions are defined on different intervals due to the location parameter. We construct a sampled version of the pdfs as follows: First we compute 100 samples of the pdf on the interval between its 5-percentile and its 95 percentile. The we embed all pdfs in the larger interval between the lowest 5 percentile and the highest 95 percentile for all pdfs under consideration. We then define 200 equally spaced sample points on the larger interval and finally construct for each pdf a 200-D vector with the square root of the pdf at these sample points by interpolation. This correponds to the Hellinger distance between two distributions~\cite{Nikulin}. Each pdf is thus characterized by a 200-D vector defined on a common reference grid.

Next we select from the 200-D vectors those $K$~samples which have the highest variance. The components in this $K$-dimensional vector are of course very similar to each other and there are other, more effective methods available, but we found this choice sufficient for this first study. 

After the feature selection we selected $M$~trials, described in the previous experiment, and the corresponding distributions for all observers. These $M\times 15$ $K$-dimensional vectors were then used to train a support vector machine (SVM) to discriminate one observer against all the other observers. After the training we selected $N$ vectors from random observers and random trials and classified them with the trained SVM. This was repeated 100~times and the results were collected in the mean recognition rates. In all experiments we used $N=5000$~vectors for classification.

\begin{figure}[htbp]
	\centering
		\includegraphics[width=0.75\columnwidth]{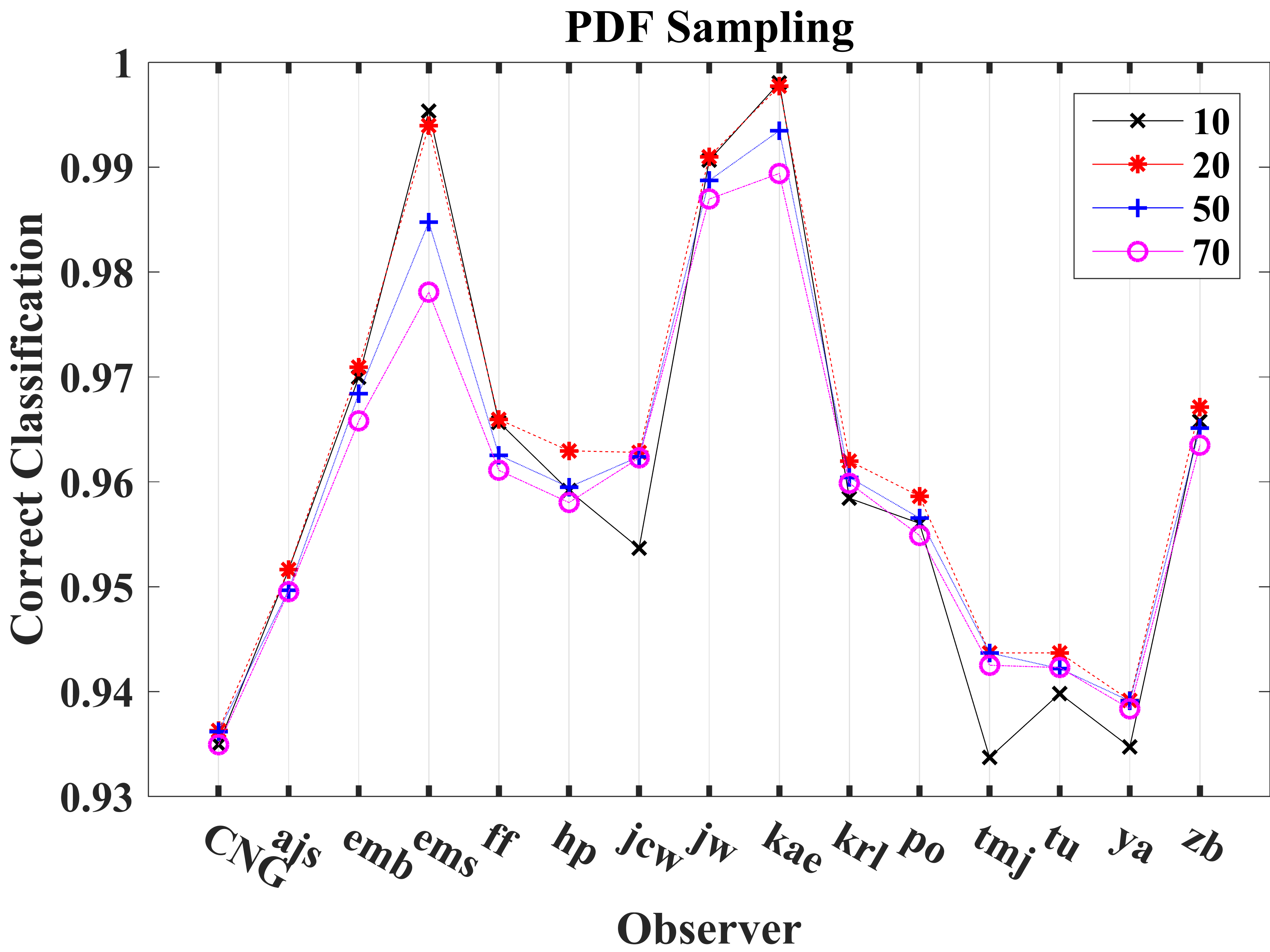}
	\caption{Recognition rates as function of the number of pdf-samples}
	\label{fig:PDFSampling}
\end{figure}

The following figures illustrate some of the results obtained using these evaluation procedure. First we investigated the influence of the number of pdf-samples used (the value of the $K$-parameter). We used the euclidean distance, 5000 trials, 50 images per trial and $M=50$ samples to train the SVM. In each classification step 5000 vectors were classified. Figure~\ref{fig:PDFSampling} shows the results obtained for~$K = \left[10, 20, 50, 70\right]$

We see that the best results are obtained using~10 or~20 samples and we therefore choose to use 20~pdf samples in the following experiments.

\begin{figure}[htbp]
	\centering
		\includegraphics[width=0.75\columnwidth]{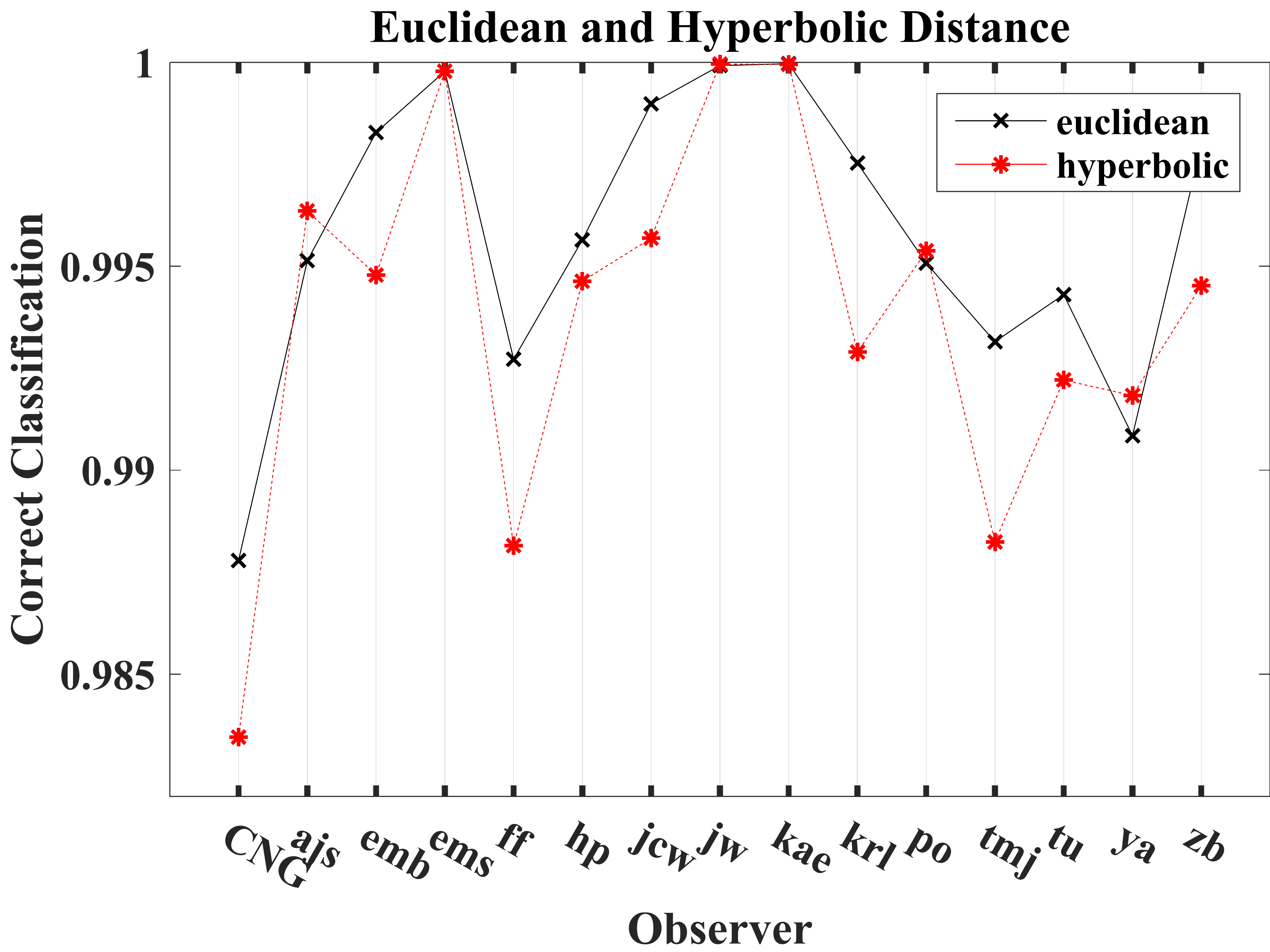}
	\caption{Euclidean vs. Hyperbolic distance, 10000 trials and 250 images per trial}
	\label{fig:HypEuc10k250}
\end{figure}

In the next experiment we investigated if the euclidean or the hyperbolic distance measure gives better classification results. We used, 10000 trials, 250 images per trial and $M=100$ samples to train the SVM. In figure~\ref{fig:HypEuc10k250} we show the results and we see that the euclidean distance measure gives in general better classification results than the hyperbolic distance (but note the very high recognition rates). 

\begin{figure}[htbp]
	\centering
		\includegraphics[width=0.75\columnwidth]{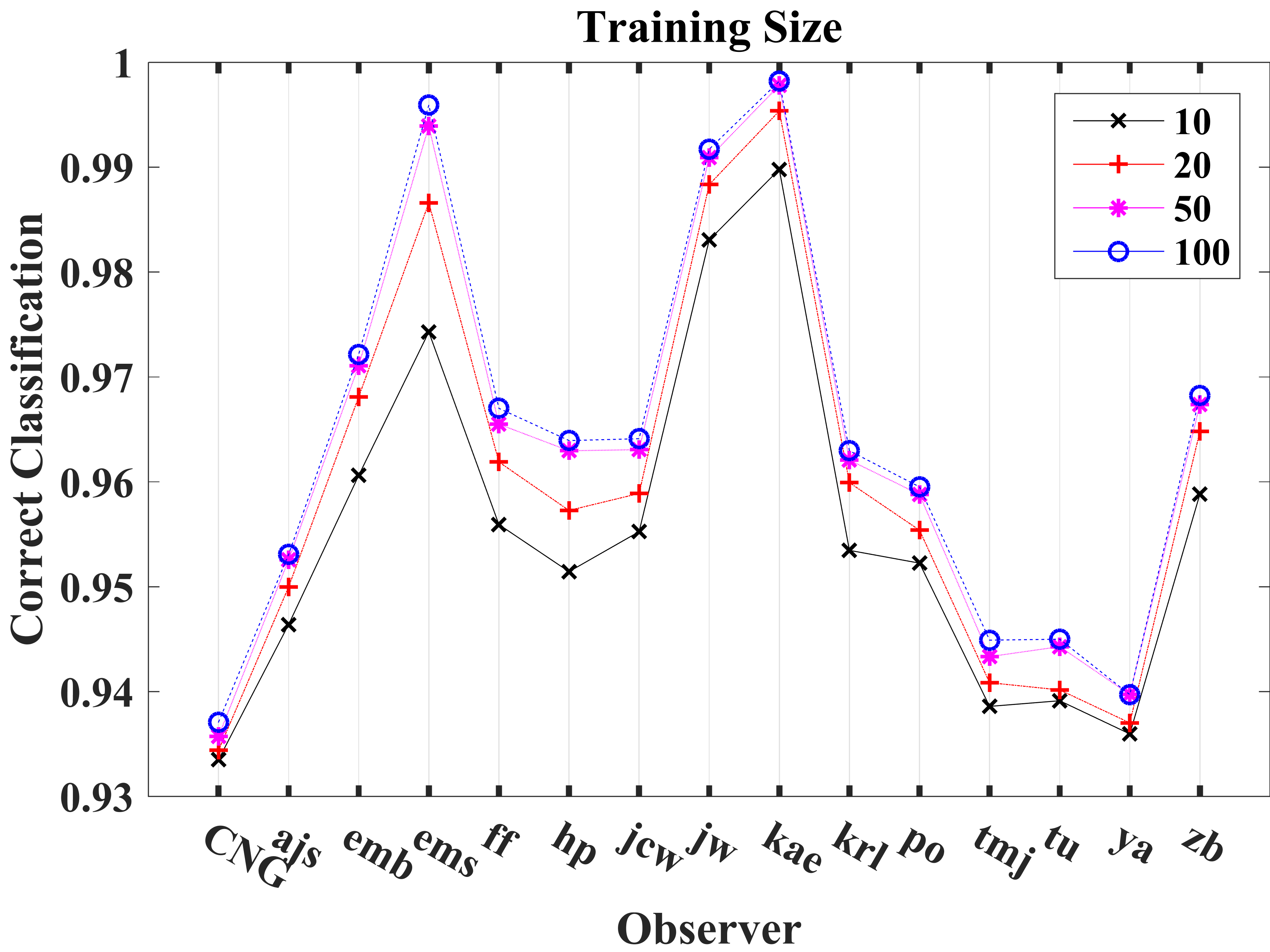}
	\caption{Influence of the size of the training set}
	\label{fig:Training500050Euc}
\end{figure}

In Fig.~\ref{fig:Training500050Euc} we illustrate the influence of the number of images used to train the SVM. We used, 5000 trials, 50 images per trial and $M=\left[10,20,50,100\right]$ samples to train the SVM. We see that~20 or~50 samples are sufficient.

In the last experiment we investigated how the number of images used in a trial influence the classification result. We used 5000~trials and 50 and 200 images per trial (corresponding to Figs.~\ref{fig:Pareto_5000_50_ksigma} and~\ref{fig:Pareto_5000_200_ksigma})
We see that the number of images used to estimate the parameters of the GPD has a great influence on the classification result, as was to be expected.

\begin{figure}[htbp]
	\centering
		\includegraphics[width=0.75\columnwidth]{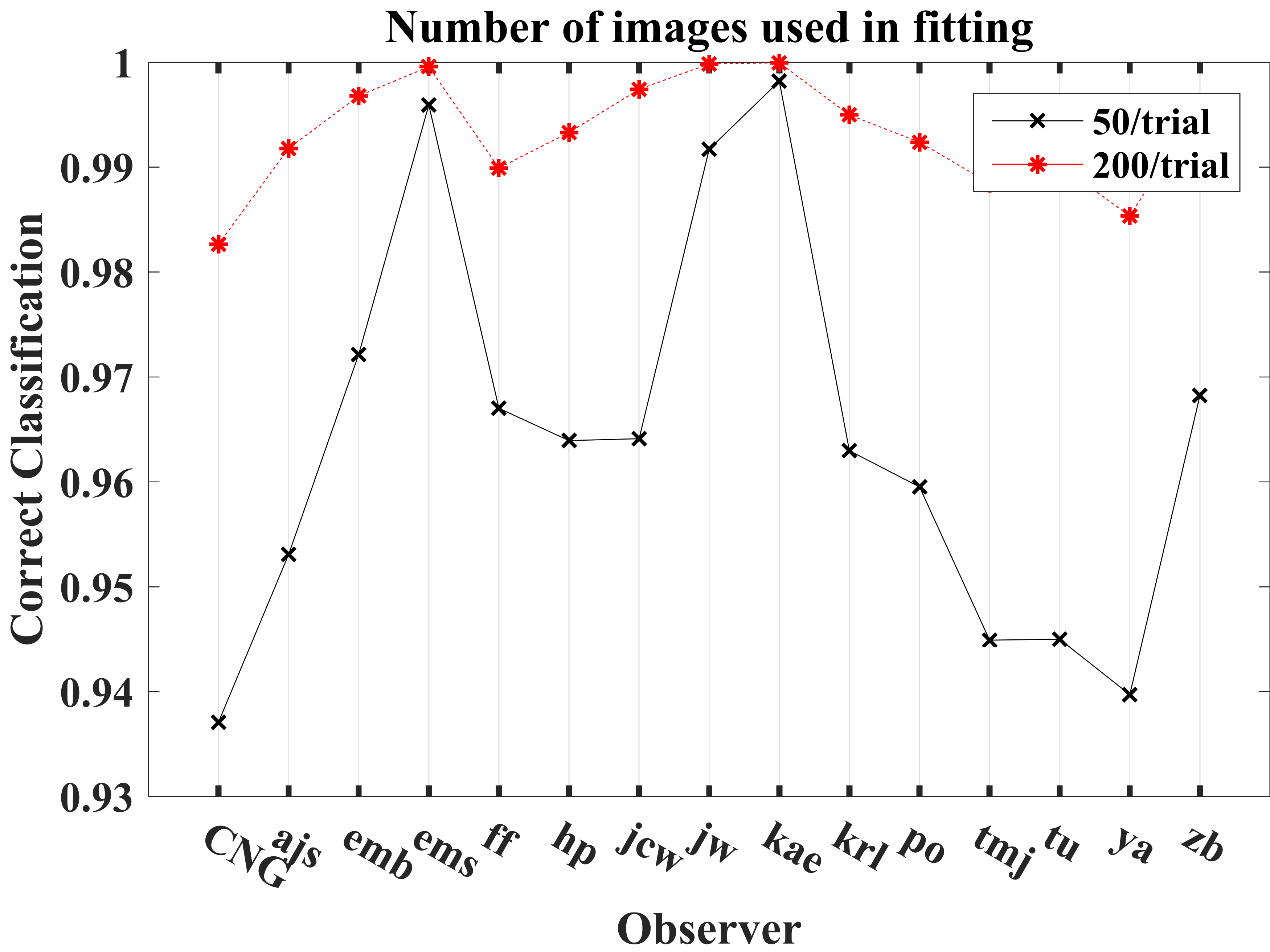}
	\caption{Varying number of images used in the distribution fitting}
	\label{fig:Euc50vs200Trial}
\end{figure}

Comparing the classification results shown in Figs.~\ref{fig:PDFSampling},~\ref{fig:HypEuc10k250},~\ref{fig:Training500050Euc} and~\ref{fig:Euc50vs200Trial} and the distribution of the parameters in Figs.~\ref{fig:Pareto_5000_50_ksigma} and~\ref{fig:Pareto_5000_200_ksigma}
we can see that observers {\it ems, jw, kae} and~{\it zb} are easy to recognize while {\it CNG} and~{\it ya} are more difficult to distinguish from their neighbors.  

\section{Discussion}

The three main topics described in the paper are (1) the introduction of the hyperbolic distance, (2) the usage of the GPD and (3) the recognition of observers based on the measurements of the saccadic eye movements. 

The application of hyperbolic geometry is attractive from a theoretical point of view since the limitation of the viewing geometry is built into the model. The experiments showed that the fitting of the GPD distribution, as measured by the R-squared values, are slightly better for the hyperbolic distance than for the euclidean distance. This is probably an effect of the relative higher importance of large distances in the hyperbolic geometry. The somewhat lower classification rates in the hyperbolic case, compared to the euclidean case, may be an indication that for a classification the eye movements in the transition phases between fixation and saccades are also important. 

In both cases, euclidean and hyperbolic, we find however that the GPD provides a very good and compact model for the statistical distribution of the step lengths. The classification experiments show that these distribution are potentially useful for identification tasks but more detailed studies with more observers are necessary to judge the potential of the method.

We also note that the special form of the GPD's make it possible to derive analytic expressions for different distances (like the Hellinger distance). We derived some of these expressions with the help of Mathematica but their final form often involves expressions based on special functions like the hypergeometric function which are costly to evaluate numerically. We therefore used the numerical implementation described above.  

No attempts were made to optimize the parameters in the different processing steps. All statistical computations (gpfit, gmdistribution,  fitcsvm, predict) were done using the tools in the Matlab 2014b Statistics Toolbox with default parameter settings.

\end{document}